\documentclass{article}

\usepackage[main, final]{neurips_2026}
\usepackage[utf8]{inputenc}
\usepackage[T1]{fontenc}
\usepackage[hidelinks]{hyperref}
\usepackage{url}
\usepackage{booktabs}
\usepackage{amsfonts}
\usepackage{amsmath}
\usepackage{amssymb}
\usepackage{nicefrac}
\usepackage{microtype}
\usepackage{xcolor}
\usepackage{graphicx}
\usepackage{multirow}
\usepackage{array}
\usepackage{enumitem}
\usepackage{tikz}
\usepackage{upquote}
\usetikzlibrary{positioning,shapes.geometric,arrows.meta,calc,fit,backgrounds}

\newif\ifdraftmarkers \draftmarkersfalse
\ifdraftmarkers
  \newcommand{\expneeded}[1]{{\color{red}\textbf{[EXP: #1]}}}
  \newcommand{\advisor}[2]{{\color{orange}\textbf{[ADVISOR\,\##1: #2]}}}
  \newcommand{\hltodo}[1]{{\color{blue}\textbf{[TODO: #1]}}}
\else
  \newcommand{\expneeded}[1]{}
  \newcommand{\advisor}[2]{}
  \newcommand{\hltodo}[1]{}
\fi
\title{Externalized CPDAG Summaries Improve LLM Causal Deduction}

\author{%
  Wentao Sun\\
  Nokia Bell Labs\\
  \texttt{wentao.sun@nokia.com}\\
  \texttt{wentao.sun@inria.fr}\\
  \AND
  Jo{\~a}o Paulo Nogueira\\
  T\'el\'ecom SudParis\\
  \texttt{joao-paulo.fontoura\_nogueira}\\
  \texttt{@telecom-sudparis.eu}\\
  \And
  Dominique Verchere\\
  Nokia Bell Labs\\
  \texttt{dominique.verchere}\\
  \texttt{@nokia-bell-labs.com}\\
  \AND
  Mathieu Acher\\
  IRISA\\
  \texttt{mathieu.acher@irisa.fr}\\
  \And
  Alonso Silva\thanks{Corresponding author.}\\
  Nokia Bell Labs\\
  \texttt{alonso.silva@nokia-bell-labs.com}\\
}

\begin{document}

\maketitle

\begin{abstract} 
Observational data rarely identify a single causal structure. Such structures are typically modeled as directed acyclic graphs (DAGs) whose arrows point from direct causes to their effects, and observed correlations and conditional independencies usually leave a whole family of DAGs that fit the data equally well. A causal claim follows from the data only if it holds in every one of them. Classical causal discovery methods such as the Peter--Clark (PC) algorithm summarize this family as a completed partially directed acyclic graph (CPDAG): an edge is directed when every compatible graph orients it the same way, and undirected when the data leave its direction open. On the Corr2Cause benchmark, large language models (LLMs) that reason in free text often lose track of this family and decide from a few local edges. We propose \emph{Structured Thinking}, a two-turn method in which the model first summarizes the CPDAG in a constrained schema and then checks the claim against that summary. Without fine-tuning or an external solver, it raises Qwen3.5-27B from $77.5$ to $86.5$ F$_1$ (mean of three seeds) over a baseline that receives explicit PC-algorithm instructions, whereas a second turn of free-text analysis does worse than that baseline. The gain holds on test splits with renamed variables or paraphrased hypotheses and on other models. Given only the graph and the claim, the model's answer flips when the graph is replaced by a valid CPDAG under which the claim has the opposite truth value. Moreover, replacing the model's graphs with reference CPDAGs on a fixed subset shows that most remaining errors arise in building the graph rather than in reading it. Making causal structure explicit can help LLMs reason more reliably, provided the intermediate representation is itself trustworthy.
\end{abstract}

\section{Introduction}
\label{sec:intro}

\begin{figure}[t]
\centering
\includegraphics[width=\linewidth]{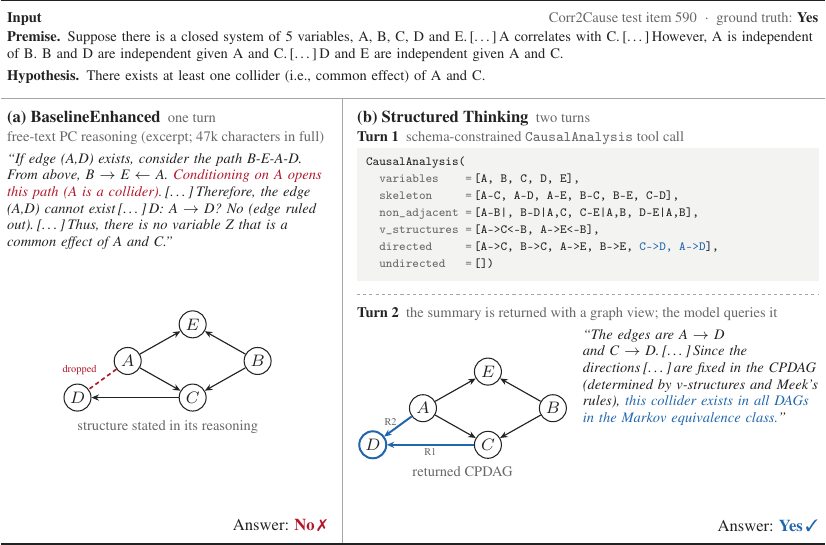}
\caption{\textbf{Free-text reasoning versus Structured Thinking on Corr2Cause test item 590} (Qwen3.5-27B, abridged verbatim outputs). With the PC-algorithm prompt alone (left), the model misapplies d-separation (red), drops the edge $A$--$D$, and finds no common effect of $A$ and $C$. Structured Thinking (right) writes the CPDAG into a schema-constrained \texttt{CausalAnalysis} call; Turn~2 reads off $A\to D\leftarrow C$ and checks that this collider holds in every DAG of the equivalence class. R1 and R2 mark orientations forced by Meek's rules.}
\label{fig:pipeline}
\vspace{-8pt}
\end{figure}

Figure~\ref{fig:pipeline} shows a test item from Corr2Cause~\citep{jin2024corr2cause}. The premise describes a closed system of five variables through the correlations and conditional independencies among them, and the hypothesis asks whether $A$ and $C$ have at least one common effect. Many causal graphs can produce the same statistics, so the hypothesis counts as true only if it holds in every one of them. Given explicit instructions for causal discovery, Qwen3.5-27B reasons for 47k characters of free text, misapplies d-separation (the graphical test for conditional independence), drops the edge $A$--$D$, and answers \textsc{No}. Asked first to write down the graph, the same model finds the collider $A\to D\leftarrow C$ and answers \textsc{Yes}.

The difficulty is that the label depends on a hidden object that the premise never states. Causal structure is modeled as a directed acyclic graph (DAG), whose arrows point from direct causes to their effects. Under standard Markov and faithfulness assumptions, the premise determines a Markov equivalence class: the set of DAGs that encode exactly the stated independencies. Classical causal discovery procedures, such as the Peter--Clark (PC) algorithm~\citep{DBLP:books/daglib/0023012,kalisch2007estimating}, summarize this class as a completed partially directed acyclic graph (CPDAG). An edge is directed in the CPDAG when every graph in the class orients it the same way, and undirected when the class contains both orientations. A hypothesis is labeled \textsc{Yes} only if it holds in every graph of the class, so each Corr2Cause question is a query on this CPDAG. Answering it in prose requires the model to hold many facts at once: which pairs of variables are adjacent, which triples form colliders, and which edge directions are forced. Free text gives these facts no fixed place, and a long trace such as the one in Figure~\ref{fig:pipeline} loses some of them along the way.

In earlier work, we found that asking a model to build a knowledge-graph style intermediate improves causal generalization on Corr2Cause~\citep{sun2025structured}. However, that study left open what drives the gain: the structure itself, the causal instructions that come with it, or simply the extra turn spent before answering.

To separate these factors, we introduce \emph{Structured Thinking}, a two-turn method that splits building the causal graph from reading it (Figure~\ref{fig:pipeline}, right). In Turn~1, the model follows PC-algorithm instructions and writes a \texttt{CausalAnalysis} object listing the variables, the skeleton, the non-adjacent pairs with their separating sets, the v-structures (colliders), and the directed and undirected edges. For Qwen-family models, decoding is constrained to this schema; API models receive the same schema through tool calling. In Turn~2, the object is returned to the model, which checks the hypothesis against the graph. The model builds and reads its own graph: nothing is fine-tuned and no solver is called.

Our main baseline, \textsc{BaselineEnhanced}, receives a detailed PC-algorithm walk-through and answers in one turn. On the full Corr2Cause test set, Structured Thinking raises Qwen3.5-27B from $77.5$ to $86.5$ F$_1$(\textsc{Yes}), averaged over three seeds, and our controls trace this gain to the graph rather than to the extra turn or the graph vocabulary. When Turn~1 writes the same PC analysis as free text, the model falls below the baseline, mostly by adding false positives. A single-turn prompt with the structured instructions but no emitted graph stays near the baseline, and removing the hard schema constraint costs much of the gain, largely because Turn~2 then often receives no usable graph. Turn~2 also relies on what the graph says. When the premise is hidden and the graph is replaced by a valid counterfactual CPDAG that reverses the true answer, Turn~2 switches to the new answer in $90\%$ of cases. Substituting reference CPDAGs for the model's own graphs removes most remaining errors on a fixed subset, mainly by supplying graphs the model failed to produce, so the main bottleneck is graph construction rather than graph reading.

The gain carries over to Qwen3.6-27B and Qwen3.5-9B and to test splits with renamed variables or paraphrased hypotheses; it is smaller for GPT-5.4-mini and Gemma-4-31B, whose baselines are already strong. Beyond Corr2Cause, a typed graph intermediate beats a matched prose intermediate on StepGame spatial deduction~\citep{shi2022stepgame,li2024stepgame}. It fails on CLUTRR kinship reasoning~\citep{sinha2019clutrr}, where even reference graphs give low accuracy because composing kinship relations over the graph is itself hard for the model.

\paragraph{Contributions.} We treat causal deduction on Corr2Cause as building and querying a CPDAG, and show that a language model can do both on its own. Using an instruction-rich baseline, a two-turn prose control, and ablations, we show that the gain cannot be explained by causal instructions or an extra turn alone. Using counterfactual graphs, reference-graph substitution, a validity audit, and two transfer tasks, we show that Turn~2 uses the content of the graph, locate most remaining errors in graph construction, and identify where the approach stops working.

\section{Background and Related Work}
\label{sec:related}

\paragraph{CPDAGs and causal deduction.} The PC algorithm~\citep{kalisch2007estimating,DBLP:books/daglib/0023012} removes edges using conditional independencies, orients v-structures, and propagates Meek rules~\citep{meek1995causal}. Its output is a CPDAG: directed edges are forced across the equivalence class, while undirected edges encode genuine ambiguity~\citep{pearl2009causality}. Corr2Cause uses this CPDAG as the hidden object behind each label. This is why the task is a better stress test than asking whether two variables are merely associated: an answer can be wrong even when the model finds a plausible DAG, if another DAG in the same equivalence class falsifies the hypothesis.

\paragraph{LLMs and structural causality.} Surveys and benchmarks consistently find structural causal reasoning harder for LLMs than commonsense causal association~\citep{kiciman2024causal,ma2025survey,zecevic2023causal,willig2022foundation,long2023causal,chi2024unveiling}. \citet{yang2024critical} argue that many causal benchmarks can be partially solved by retrieving familiar domain knowledge; Corr2Cause is useful precisely because its labels are generated from independence constraints and PC-algorithm semantics. Complementary benchmarks such as CLadder, CLEAR-3K, and CausalARC evaluate questions across Pearl's ladder of causation, natural-language causal explanation, and SCM-world reasoning rather than CPDAG deduction~\citep{jin2023cladder,liu2026clear3k,maasch2025causalarc}. Prompting strategies improve performance~\citep{sgouritsa2024prompting,liu2025eliciting,bagheri2024c2p}, but still leave a gap to algorithmic ground truth, especially as the number of variables grows.

\paragraph{Structured intermediates.} The closest precursor to this work builds a knowledge-graph style intermediate for Corr2Cause and shows that explicit structure can improve causal generalization~\citep{sun2025structured}. We make the intermediate more specific: it is a CPDAG summary, emitted under regex- or JSON-schema-constrained decoding~\citep{willard2023outlines,dong2024xgrammar}, and compared against a strong direct-answer PC baseline. Concurrent modular prompting decomposes Corr2Cause into prose sub-answers~\citep{kadziolka2025pieces}; C2P similarly uses causal prompting steps but is not a constrained CPDAG-state method~\citep{bagheri2024c2p}. A prose sub-answer can be locally plausible but hard to audit; a typed graph has explicit fields whose content can be removed, scrambled, or queried.

\paragraph{Faithfulness of reasoning traces.} Recent work questions whether free-form CoT is causally used by the final answer, using pre-CoT probes, causal-bypass diagnostics, causal sufficiency/necessity tests, and judge benchmarks for process faithfulness~\citep{cox2025posthoc,sathyanarayanan2026bypass,yu2025sufficiency,mittal2026c2faith,fu2025injecting}. Our probes are behavioral rather than activation-level: after Turn~1 emits a graph, we intervene on graph fields and measure Turn~2 degradation. They therefore test content dependence of the emitted state, not full mechanistic faithfulness of every token.

\paragraph{Tool use, programs, and positioning.} Tool-augmented language models~\citep{parisi2022talm,yao2023react,schick2023toolformer} usually call external functions, while program-aided prompting~\citep{gao2023pal,chen2023pot} offloads computation to code. Structured Thinking instead uses the tool interface to record the model's own schema-constrained analysis; no symbolic causal solver supplies the answer. Our contribution is a constrained CPDAG-summary intermediate, evaluated against direct-answer and two-turn prose PC baselines and tested through interventions on graph content.

\section{Method}
\label{sec:method}

\subsection{Task}
\label{sec:task}

An instance has a premise $P$ listing marginal dependencies and conditional independencies $X_i \perp X_j \mid Z$, plus a hypothesis $H$ such as ``$X_a$ is a parent of $X_b$.'' Under the faithfulness and Markov assumptions, $P$ identifies a CPDAG $\mathcal{G}$ and its Markov equivalence class $\mathrm{MEC}(\mathcal{G})$. The label is
\[
y(P, H) \;=\; \textsc{Yes} \iff \forall D \in \mathrm{MEC}(\mathcal{G}): H \text{ holds in } D,
\]
i.e.\ the hypothesis must survive \emph{all} valid DAG completions. We report F$_1$ on the positive class as the primary metric because the ID split is imbalanced toward \textsc{No} ($180/1162$ positives), and report accuracy as a secondary metric.

\subsection{Structured Thinking pipeline}
\label{sec:pipeline}

Structured Thinking is a two-turn conversation. In Turn~1, the system prompt describes the PC algorithm and requires a \texttt{CausalAnalysis} object with fields \texttt{variables}, \texttt{skeleton}, \texttt{non\_adjacent}, \texttt{v\_structures}, \texttt{directed}, and \texttt{undirected}. For Qwen-family open-weight models, decoding is constrained by a schema-derived regex; for API models, the same schema is enforced through JSON-schema tool calling. Gemma uses the same structured interface without hard constrained decoding because the available backend was unreliable for that tokenizer. Thinking tokens, when used, appear before the constrained region. In Turn~2, the emitted object $g$ is returned as a tool-result message with a compact DOT view, and the model answers $H$ by checking adjacency, forced orientations, and directed reachability over all valid completions. The final \textsc{Yes}/\textsc{No} is extracted by regex.

The schema guarantees parseable fields; acyclicity, Meek closure, and CPDAG consistency are left to the model. We therefore distinguish three objects: a \emph{candidate CPDAG summary} is what the model emits; a \emph{graph-validated CPDAG} is a candidate that passes our validator (\S\ref{sec:error-localization}); and a \emph{reference CPDAG} is built from the premise by a symbolic solver. The design pins both the intermediate \emph{type} (a CPDAG summary rather than prose) and the Turn~2 \emph{target} (a graph query rather than a restatement of $P$). We ablate the PC prompt, constrained decoding, and thinking tokens, and probe premise visibility and graph content, in \S\ref{sec:ablation}--\S\ref{sec:probes}; full prompt and schema text appears in Appendix~\ref{app:prompts}--\ref{app:schema}.

\paragraph{Why this intermediate is useful.} A CPDAG separates three decisions that free text tends to entangle: which pairs are adjacent, which triples are compelled colliders, and which orientations are forced across the equivalence class. Turn~2 can inspect this structure directly rather than rediscover it from the premise (Figure~\ref{fig:pipeline}); the advantage is a better state representation, not more verbosity.

\paragraph{Constrained decoding.} The constraint is applied to the \emph{analysis object}, not to the free reasoning tokens that precede it. For Qwen-family models, the regex wraps a \texttt{<think>} block followed by a \texttt{<tool\_call>} envelope. Removing the hard constraint keeps the tool-call template as a soft instruction, so the ablation asks whether syntactic enforcement itself matters beyond reminding the model to be structured.

\subsection{Baselines and controls}
\label{sec:baseline}

The principal baseline, \textsc{BaselineEnhanced}, is a strong direct-answer PC baseline. It receives an explicit PC-algorithm walk-through in a single turn and answers directly. The prompt covers skeleton removal from conditional independencies, collider orientation, Meek-rule propagation, closed-world treatment of unstated independencies, and the requirement that a positive hypothesis hold in every DAG in the Markov equivalence class. This comparison controls for giving the direct-answer model an explicit PC-algorithm scaffold.

As a prose-state control, we add \textsc{TwoTurn-Prose-PC}. Turn~1 uses the same detailed PC scaffold as \textsc{BaselineEnhanced} but asks for free-text analysis while withholding the final yes/no answer; Turn~2 returns that analysis and requests the answer. This control uses no tool call and no schema-constrained typed graph object. It tests whether the same PC scaffold plus a two-turn prose intermediate suffices.

We also include a single-turn structured-prompt diagnostic for Qwen3.6-27B, \textsc{SingleTurn-StructPrompt}. This condition reuses the structured CPDAG-oriented instructions but does not require a tool call or schema-constrained graph output before answering. It tests whether merely mentioning the right graph vocabulary can replace writing out the graph itself.

\section{Main Results}
\label{sec:results}

Our experiments answer four questions. \emph{Effectiveness}: does a typed intermediate improve over a strong direct PC-instruction baseline and over a PC-scaffolded prose intermediate? \emph{Robustness}: does the gain hold across seeds, models, perturbed wording, and graph size? \emph{Validity and use}: how often is the emitted state a valid CPDAG, and does Turn~2 depend on its content? \emph{Scope}: where does the recipe transfer, and where does it fail?

\subsection{Setup}
\label{sec:setup}

\paragraph{Models and decoding.} Unless noted, open-weight models (\texttt{Qwen3.5-27B}, \texttt{Qwen3.6-27B}, \texttt{Qwen3.5-9B}, \texttt{Gemma-4-31B-it}) run under vLLM~\citep{kwon2023vllm} with $T{=}0.6$, top-$p{=}0.95$, \texttt{max\_model\_len}$=\!32768$, and fixed seed; \texttt{gpt-5.4-mini} uses the OpenAI Responses API with \texttt{reasoning\_effort="high"}; model identifiers are listed in Appendix~\ref{app:assets}. Qwen-family Structured Thinking runs use schema-derived regex constraints via \texttt{xgrammar}~\citep{dong2024xgrammar}; direct baselines and \textsc{TwoTurn-Prose-PC} do not emit a constrained CPDAG object. Gemma Structured is reported without constrained decoding because available backends were unreliable for that tokenizer.

\paragraph{Data and protocol.} ID is the $n{=}1162$ Corr2Cause test split. We also use Corr2Cause's two perturbed test splits ($n{=}2246$ each): \emph{Refactor-OOD} renames the variables, and \emph{Paraphrase-OOD} rewrites the hypotheses. The headline Qwen3.5-27B ID comparison is a seed-matched rerun on seeds $42$, $52$, and $62$ with the dataset hash, model revision, and decoding settings pinned; for each seed we report an exact McNemar test, and across seeds the mean and sample standard deviation. All remaining rows are single runs, with the original Qwen3.5-27B run retained for the controls, ablations, and probes.

\subsection{Results across seeds, splits, and models}

\begin{table}[t]
\caption{\textbf{Main results on Corr2Cause}. F$_1$(\textsc{Yes}) and accuracy (\%). Top: seed-matched rerun, mean $\pm$ sample SD over three seeds (per seed in Table~\ref{tab:seed-robustness}). Bottom: single runs. $\Delta$ is the F$_1$ change relative to the one-turn PC-instruction baseline for the same model and split. CD marks hard schema-constrained decoding of the Turn-1 graph.}
\label{tab:main}
\centering
\small
\setlength{\tabcolsep}{3.3pt}
\begin{tabular}{llccccr}
\toprule
\textbf{Model} & \textbf{Method} & \textbf{Split} & \textbf{CD} & \textbf{F$_1$(Yes)} & \textbf{Acc.} & \textbf{$\Delta$F$_1$} \\
\midrule
\multirow{2}{*}{\shortstack[l]{Qwen3.5-27B\\(3 seeds)}}
& BaselineEnhanced & ID & -- & $77.48 \pm 3.23$ & $93.55 \pm 0.95$ & --- \\
& Structured Thinking & ID & \checkmark & $\mathbf{86.48} \pm 1.36$ & $\mathbf{95.81} \pm 0.41$ & $+9.00$ \\
\midrule
\multirow{7}{*}{Qwen3.5-27B}
& BaselineEnhanced & ID & -- & $73.01$ & $92.43$ & --- \\
& TwoTurn-Prose-PC & ID & -- & $67.55$ & $89.41$ & $-5.46$ \\
& Structured Thinking & ID & \checkmark & $\mathbf{86.36}$ & $\mathbf{95.87}$ & $+13.35$ \\
& BaselineEnhanced & Refactor-OOD & -- & $72.33$ & $92.61$ & --- \\
& Structured Thinking & Refactor-OOD & \checkmark & $\mathbf{82.79}$ & $\mathbf{95.06}$ & $+10.46$ \\
& BaselineEnhanced & Paraphrase-OOD & -- & $75.43$ & $92.92$ & --- \\
& Structured Thinking & Paraphrase-OOD & \checkmark & $\mathbf{84.21}$ & $\mathbf{94.92}$ & $+8.79$ \\
\midrule
\multirow{3}{*}{Qwen3.6-27B}
& BaselineEnhanced & ID & -- & $75.92$ & $92.69$ & --- \\
& SingleTurn-StructPrompt & ID & -- & $74.53$ & $91.82$ & $-1.39$ \\
& Structured Thinking & ID & \checkmark & $\mathbf{85.71}$ & $\mathbf{95.52}$ & $+9.79$ \\
\midrule
\multirow{2}{*}{Qwen3.5-9B}
& BaselineEnhanced & ID & -- & $39.18$ & $84.77$ & --- \\
& Structured Thinking & ID & \checkmark & $\mathbf{59.56}$ & $\mathbf{88.90}$ & $+20.38$ \\
\midrule
\multirow{2}{*}{GPT-5.4-mini}
& BaselineEnhanced & ID & -- & $85.39$ & $95.61$ & --- \\
& Structured Thinking & ID & \checkmark & $\mathbf{88.58}$ & $\mathbf{96.47}$ & $+3.19$ \\
\midrule
\multicolumn{7}{l}{\emph{Additional observation: no hard constrained decoding}} \\
\multirow{2}{*}{Gemma-4-31B}
& BaselineEnhanced & ID & -- & $85.38$ & $95.70$ & --- \\
& Structured Thinking & ID & $\times$ & $\mathbf{87.39}$ & $\mathbf{96.13}$ & $+2.01$ \\
\bottomrule
\end{tabular}
\end{table}

Across three matched seeds, Structured Thinking raises Qwen3.5-27B from $77.48$ to $86.48$ F$_1$(\textsc{Yes}) ($+9.00 \pm 4.56$\,pp) and from $93.55\%$ to $95.81\%$ accuracy (Table~\ref{tab:main}). All three gains are positive: $+12.84$, $+3.96$, and $+10.20$\,pp on seeds $42$, $52$, and $62$. Across seeds, Structured Thinking fixes $141$ baseline errors and introduces $62$ new ones; fixes outnumber new errors on every seed. The exact McNemar test is significant on seeds $42$ and $62$ ($p=4.2\times10^{-6}$ and $1.1\times10^{-3}$) but not on seed $52$ ($p=0.22$), which has the strongest baseline. The direction of the gain is stable; its size varies from run to run.

The gain comes mainly from recall. Per seed, recall rises from $70.0$, $75.6$, and $69.4\%$ to $86.7$, $84.4$, and $88.3\%$, while precision moves from $79.8$, $87.2$, and $86.2\%$ to $88.1$, $85.4$, and $86.0\%$. On average, recall rises by $14.8$\,pp and precision by $2.1$\,pp. Structured Thinking is also more stable across seeds: $48$ of $1162$ items change prediction across the three seeds, against $111$ for the baseline, whose unstable items all have five or more variables and are mostly positives.

\textsc{TwoTurn-Prose-PC} reaches only $67.55$ F$_1$, below both the one-turn baseline and Structured Thinking. It has higher recall than the one-turn baseline, but at the cost of many more false positives ($71$ versus $27$; Appendix Table~\ref{tab:cm}): with the same PC scaffold, a schema-free prose intermediate mostly shifts answers toward \textsc{Yes}. In the same run, Structured Thinking reduces both false negatives ($61{\to}28$) and false positives ($27{\to}20$).

The same pattern holds beyond one Qwen release. On Qwen3.6-27B, Structured Thinking improves from $75.92$ to $85.71$ F$_1$ ($+9.79$\,pp). A single-turn structured-prompt diagnostic reaches $74.53$, so merely giving CPDAG-oriented wording without requiring an external graph object does not explain the gain.

The gain also persists when the variables are renamed ($+10.46$\,pp) and when the hypotheses are paraphrased ($+8.79$\,pp; exact McNemar $p=3.5\times10^{-6}$ and $7.7\times10^{-4}$). The remaining rows compare each model with its own baseline. Structured Thinking improves Qwen3.5-9B by $+20.38$\,pp and GPT-5.4-mini by $+3.19$\,pp. Gemma-4-31B, which runs without hard constrained decoding, gains $+2.01$\,pp. Gains are smaller when the direct baseline is already strong, as it is for GPT-5.4-mini and Gemma (about $85$ F$_1$).

\subsection{Gains concentrate on larger graphs}

\begin{figure}[t]
\centering
\includegraphics[width=\linewidth]{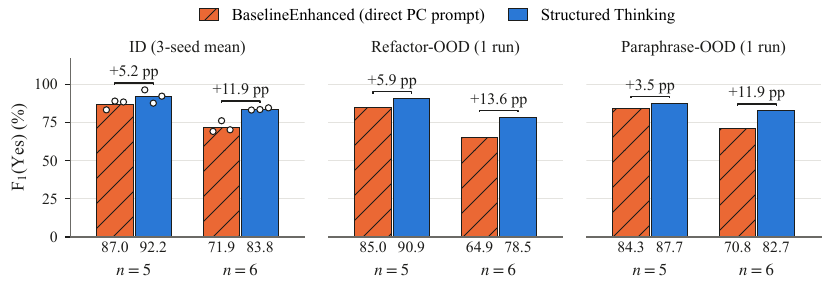}
\caption{\textbf{F$_1$(\textsc{Yes}) by variable count (Qwen3.5-27B).} ID bars are three-seed means, with dots for seeds $42$, $52$, and $62$ from left to right; the perturbed splits are single runs. Brackets give the gain in pp. Bins with $n{\le}4$ contain at most three positives and are omitted.}
\label{fig:complexity}
\end{figure}

Figure~\ref{fig:complexity} shows that the gain is larger in the hardest bin: on ID, the three-seed gain grows from $+5.2$\,pp at $n{=}5$ to $+11.9$\,pp at $n{=}6$, and the perturbed splits show the same pattern. At $n{=}6$, Structured Thinking is also nearly seed-invariant ($83.2$--$84.7$ F$_1$), while the baseline ranges from $69.2$ to $76.2$.

Pilot stress tests on larger graphs from other graph families ($n{=}7$ and $8$; Appendix~\ref{app:larger-graph}) show that the advantage extends beyond the original $n{\le}6$ regime. On Qwen3.5-27B, Structured Thinking improves Random $n{=}7$ from $28.42$ to $91.87$ F$_1$, ALARM-derived $n{=}7$ from $15.38$ to $88.73$, and ASIA-derived $n{=}8$ from $9.30$ to $60.00$. In all three settings the direct baseline's recall is low ($9$--$18\%$); for named-network-derived rows, source names and semantic variable names are withheld from the model.

\section{Ablations}
\label{sec:ablation}

\begin{table}[t]
\caption{\textbf{Ablations on Qwen3.5-27B}. $n{=}1162$ ID split. Removing the PC-algorithm prompt collapses the method almost entirely; removing thinking or constrained decoding each causes a substantial drop.}
\label{tab:ablation}
\centering
\small
\begin{tabular}{lccr}
\toprule
\textbf{Configuration} & \textbf{F$_1$(Yes)} & \textbf{Acc.} & \textbf{$\Delta$ vs.\ full} \\
\midrule
Structured Thinking (full)          & $\mathbf{86.36}$ & $95.87$ & --- \\
\hspace{1em}--thinking              & $71.95$ & $92.08$ & $-14.41$ \\
\hspace{1em}--constrained decoding  & $76.88$ & $93.37$ & $-9.48$ \\
\hspace{1em}--PC-algorithm prompt   & $36.21$ & $87.26$ & $-50.15$ \\
\midrule
BaselineEnhanced                    & $73.01$ & $92.43$ & $-13.35$ \\
\bottomrule
\end{tabular}
\end{table}

We stress-test the Qwen3.5-27B gain by removing one design choice at a time (Table~\ref{tab:ablation}). The PC-algorithm prompt matters most: replacing it with a minimal ``produce a causal analysis'' instruction drops F$_1$ to $36.2$, far below baseline. On top of that prompt, removing constrained decoding costs $9.5$\,pp and removing thinking costs $14.4$\,pp at 27\,B.

In the constrained-decoding ablation the model still sees the tool-call syntax; only the hard regular-language constraint is removed. Our validity audit shows what the constraint buys. In matched full-split runs, it raises the share of instances with a usable graph from $74.5\%$ to $97.2\%$, while the share of valid CPDAGs among usable graphs stays similar ($81.1\%$ without and $79.0\%$ with the constraint). The clearest effect of the constraint is thus that Turn~2 reliably receives a graph.

A smaller-model factorial is reported in Appendix~\ref{app:scale-thinking}: at 9\,B, thinking becomes harmful while constrained decoding remains helpful, suggesting that explicit reasoning traces should be calibrated by model size.

\section{Probes: is the emitted graph actually used?}
\label{sec:probes}

We test whether Turn~2 uses the content of the graph with validity-preserving counterfactuals, then substitute reference CPDAGs to separate building the graph from reading it (Table~\ref{tab:probes}; Qwen3.5-27B, ID split).

\begin{table}[t]
\caption{\textbf{Turn-2 probes on Qwen3.5-27B, ID split.} (a) Graph-only counterfactual probe: Turn~2 receives one of three graph-validated CPDAGs and the hypothesis, without the premise; flips are changes from the answer given with the reference CPDAG (all values in \%). (b) Accuracy (\%) on $128$ frozen instances (seed $62$) when Turn~2 receives the model's own candidate graph or the reference CPDAG, with or without the premise; a missing graph counts as an error.}
\label{tab:probes}
\centering
\small
\begin{tabular}{lccc}
\multicolumn{4}{l}{\textbf{(a)} Counterfactual probe, $192$ triplets} \\
\toprule
\textbf{Graph given to Turn~2} & \textbf{Answer flips} & \textbf{Flips toward graph truth} & \textbf{Acc.\ vs.\ graph truth} \\
\midrule
Reference CPDAG & --- & --- & $96.4$ \\
Serialization-only control & $1.0$ & --- & $95.3$ \\
Valid counterfactual CPDAG & $90.1$ & $90.1$ & $93.8$ \\
\bottomrule
\end{tabular}

\vspace{6pt}
\begin{tabular}{lcc}
\multicolumn{3}{l}{\textbf{(b)} Graph construction versus graph reading, $n{=}128$} \\
\toprule
\textbf{Graph given to Turn~2} & \textbf{Without premise} & \textbf{With premise} \\
\midrule
Candidate graph (model's own) & $89.06$ & $92.97$ \\
Reference CPDAG & $94.53$ & $96.88$ \\
\bottomrule
\end{tabular}
\end{table}

\paragraph{Validity-preserving counterfactuals.} For $192$ frozen ID instances, we pair the reference CPDAG with a counterfactual: another graph-validated CPDAG over the same variables under which the hypothesis has the opposite truth value. A serialization-only control keeps the reference graph and only permutes the order of its edges and field entries. All three graphs are valid CPDAGs, so this probe avoids the confound of graph scrambles, which can leave the space of valid CPDAGs (Appendix~\ref{app:scramble}). Given only the graph and the hypothesis, Turn~2 follows the counterfactual: the answer flips toward the new graph's truth in $173$ of $192$ triplets ($90.1\%$), against $2$ of $192$ ($1.0\%$) for the serialization control (exact McNemar $p=6.7\times10^{-52}$). When the original premise is also shown, the counterfactual graph contradicts it, and the effect shrinks but stays well above control ($28.1\%$ target-consistent flips versus $3.1\%$ control changes). Turn~2 thus weighs the premise and the graph together, and the graph's content, not its formatting, moves the answer.

\paragraph{Building the graph versus reading it.} Table~\ref{tab:probes}b substitutes solver-built reference CPDAGs for the model's own graphs. With the premise present, reference graphs raise accuracy from $92.97\%$ to $96.88\%$. The gain comes from the $6$ instances where the model emitted no usable graph: on the $122$ instances with a usable candidate graph, both arms give the same answer on every instance. Replications on seeds $42$ and $52$ agree: reference CPDAGs reduce the errors from $15$ to $5$ and from $18$ to $4$, chiefly by supplying missing graphs. On this subset, most errors therefore arise in Turn~1, chiefly as missing graphs, while Turn~2 reads a usable graph reliably but not perfectly: with the reference CPDAG and the premise, it still errs on $4$ of $128$ instances. A full-split analysis agrees: among the $945$ graph-validated CPDAGs of one full run ($81.3\%$ of instances), the graph itself implies the correct label in $98.4\%$ of cases, and Turn~2 agrees with its graph in $98.7\%$.

\section{Analysis and Practitioner Findings}
\label{sec:analysis}

\subsection{Graph validity and error localization}
\label{sec:error-localization}

\paragraph{Graph validity.} Our validator checks the graph-theoretic conditions of a CPDAG: consistent fields without duplicates, a DAG extension whose colliders are exactly the emitted v-structures, and agreement of the emitted directed and undirected edges with the CPDAG of that equivalence class, obtained by orienting the colliders and closing under Meek's rules (Appendix~\ref{app:validator}). It does not check separating-set tokens or consistency with the premise. On two full ID runs, $76.8\%$ and $81.3\%$ of test instances yield a graph-validated CPDAG; without hard constraints the rate falls to $60.4\%$, mostly because fewer graphs are produced (\S\ref{sec:ablation}).

\begin{table}[t]
\caption{\textbf{Automatic CPDAG-field audit} for Structured Thinking on full ID and Refactor-OOD splits. Graph-consistency violations are graph-semantic failures among schema-valid outputs, including field conflicts, v-structure inconsistencies, missing closure, or no valid DAG extension.}
\label{tab:cpdag-audit}
\centering
\small
\begin{tabular}{lcc}
\toprule
\textbf{Audit metric} & \textbf{ID} & \textbf{Refactor-OOD} \\
\midrule
Skeleton F$_1$ & $0.960$ & $0.959$ \\
Directed-edge F$_1$ & $0.896$ & $0.884$ \\
V-structure F$_1$ & $0.923$ & $0.920$ \\
Exact CPDAG match & $75.9\%$ & $74.4\%$ \\
Graph-consistency violations & $18.4\%$ & $20.7\%$ \\
Acc. when graph exact & $97.2\%$ & $96.6\%$ \\
Acc. with consistency violation & $91.1\%$ & $90.1\%$ \\
\bottomrule
\end{tabular}
\end{table}

\paragraph{Automatic CPDAG-field audit.}
The emitted fields agree closely with the reference CPDAG (Table~\ref{tab:cpdag-audit}): on the full ID and Refactor-OOD splits, skeleton F$_1$ is $0.96$, directed-edge F$_1$ is $0.88$--$0.90$, and $75.9\%$ and $74.4\%$ of graphs match the reference exactly. The audit also shows the remaining gap between schema validity and graph-semantic validity: graph-consistency violations occur in $18.4\%$ of ID outputs and $20.7\%$ of Refactor-OOD outputs, with Meek-closure violations in $5.1\%$ and $6.0\%$, respectively. These failures are not uniformly catastrophic: many are local or irrelevant to the queried hypothesis, and Turn~2 still sees the premise and hypothesis. Nevertheless, graph quality predicts downstream correctness. ID accuracy drops from $97.2\%$ when the emitted CPDAG exactly matches the reference to $93.9\%$ for consistent but non-exact graphs and $91.1\%$ for summaries with graph-consistency violations; Refactor-OOD shows the same ordering ($96.6\%$, $92.8\%$, $90.1\%$).

A 32-case manual diagnostic in Appendix~\ref{app:failures} gives the same qualitative picture: many residual errors are already visible in the emitted graph, while others come from Turn-2 readout or MEC-ambiguity handling.

\subsection{What the controls establish}
\label{sec:alternative-explanations}

None of these controls reproduces the gain. \textsc{BaselineEnhanced} supplies explicit PC instructions, \textsc{TwoTurn-Prose-PC} adds a prose intermediate, and \textsc{SingleTurn-StructPrompt} adds CPDAG-oriented wording without an emitted graph; all score well below Structured Thinking (\S\ref{sec:results}). The probes show that Turn~2 follows the content of the graph and is largely insensitive to its serialization (\S\ref{sec:probes}).

\subsection{Transfer beyond Corr2Cause}
\label{sec:transfer}

Beyond CPDAG queries, we compare a typed graph intermediate with a matched two-turn prose intermediate on two GPT-5.6 snapshots (Appendix~\ref{app:transfer}). On StepGame spatial deduction~\citep{shi2022stepgame,li2024stepgame}, the typed graph wins on both snapshots: $83.1\%$ versus $70.6\%$ accuracy on \texttt{gpt-5.6-terra} and $71.4\%$ versus $19.4\%$ on \texttt{gpt-5.6-luna} ($n{=}360$ each; exact McNemar $p\le1.9\times10^{-6}$). Answering directly from the story is near ceiling on this task ($99.7\%$ and $99.4\%$) and remains stronger, so StepGame tests typed against prose externalization, not the value of an intermediate over direct answering. On CLUTRR kinship reasoning~\citep{sinha2019clutrr}, the recipe fails: typed graphs fall well below the prose intermediate ($17.8\%$ and $22.2\%$ versus $70.6\%$ and $55.0\%$), and even reference graphs reach only $34\%$, so our graph representation and readout do not fit that task.

\subsection{Cost, reproducibility, and deployment}
\label{sec:cost}

Structured Thinking requires two model calls and, for Qwen-family local runs, a schema-constrained decoder, but no fine-tuning, learned verifier, or symbolic causal solver at inference time. The token footprint is roughly $2\times$ the one-turn baseline and wall-clock is about $1.5\times$ in batched vLLM runs. API-backed reasoning models require care: on GPT-5.4-mini, strict tool forcing compressed reasoning before the tool call and \texttt{tool\_choice="auto"} restored performance, whereas on a current GPT-5.6 snapshot forcing made no detectable difference (Appendix~\ref{app:api-config}). The headline comparison uses the open-weight Qwen3.5-27B; code, data, prompts, schemas, and metadata are publicly released (Appendix~\ref{app:repro}).

\section{Limitations}
\label{sec:limitations}

Our causal claims are scoped to CPDAG-style deduction from conditional independence statements. StepGame shows that typed externalization can beat a prose intermediate on another structured-deduction task, and CLUTRR shows a task where our graph-plus-readout pipeline fails. Transfer to SCM, interventional, and counterfactual queries, as in CLadder~\citep{jin2023cladder}, or to real-world causal claims is not established.

The headline Qwen3.5-27B comparison rests on three seeds. The gain is positive on each, but its size varies from $+4.0$ to $+12.8$\,pp and is not individually significant on seed $52$; with three seed-level observations, we report the spread descriptively. The perturbed splits, larger-graph stress tests, other models, and controls are single runs and do not establish robustness across seeds.

The evidence is uneven across the two turns. For graph construction, the validity audit covers full test splits, but $19$--$23\%$ of test instances do not yield a graph-validated CPDAG, and the validator does not check separating-set tokens or consistency with the premise. For graph reading, the reference-CPDAG and counterfactual probes cover frozen subsets of $128$ and $192$ instances.

Finally, schema constraints, graph-field names, tool-result formatting, and Turn-2 query instructions are bundled in the method, and our controls do not separate their contributions; the baseline's collider rule also omits the separating-set condition stated in the structured prompt (Appendix~\ref{app:prompts:baseline}). Lightly structured prose intermediates would sharpen this decomposition.

\section{Conclusion}
\label{sec:conclusion}

Structured Thinking asks a language model to write down the causal graph behind a Corr2Cause question before answering it, and on Qwen3.5-27B it raises F$_1$ by $9.0$ points on average over a baseline that already receives the PC algorithm in its prompt. Our controls show where this gain comes from. Causal instructions and an extra turn are not enough on their own: the instructions are already in the baseline, CPDAG vocabulary without an emitted graph does not improve on it, and a second turn of free-text analysis makes the model worse. The improvement appears once the intermediate is a graph in a fixed schema, and replacing that graph with a different valid one changes the answer accordingly.

Separating graph construction from graph reading also shows where to improve next. Given a valid CPDAG, the model nearly always reads it correctly, so most of its errors start earlier: a sizable share of its own graphs are missing or fail our validity check, and missing graphs account for most errors on our fixed subset. Pairing constrained decoding with a graph validator, or letting the model repair its graph before Turn~2 reads it, are natural next steps.

The same separation marks where the approach stops working. It helps when the answer can be read off the graph; on CLUTRR, composing kinship relations over the graph remains hard for the model even with reference graphs, and a prose intermediate does better. More generally, an explicit causal structure makes a model's reasoning more reliable when the model can build that structure correctly and the answer follows from it directly.

\begin{ack}
We thank the reviewers and the area chair for their constructive feedback. This work was funded by Nokia Bell Labs. Wentao Sun's doctoral research is supported by an ANRT CIFRE fellowship with Nokia Bell Labs. Competing interests: Wentao Sun, Dominique Verchere, and Alonso Silva are employees of Nokia Bell Labs; the authors declare no other competing interests.
\end{ack}

\bibliographystyle{plainnat}
\bibliography{reference}

\appendix

\section{Additional diagnostics}
\label{app:diagnostics}

\begin{table}[ht]
\caption{\textbf{Confusion matrices, Qwen3.5-27B ID split ($n{=}1162$), single run of Table~\ref{tab:main}.} Structured Thinking reduces both false negatives ($61 \to 28$) and false positives ($27 \to 20$). The PC-scaffolded prose control increases recall but introduces many false positives.}
\label{tab:cm}
\centering
\small
\begin{tabular}{llcccccc}
\toprule
 & & \multicolumn{2}{c}{\textbf{Predicted}} & & \textbf{TP} & \textbf{FN} & \textbf{FP} \\
\cmidrule{3-4}
\textbf{Method} & \textbf{Actual} & \textsc{No} & \textsc{Yes} & & & & \\
\midrule
\multirow{2}{*}{BaselineEnhanced}
& \textsc{No}  & $955$ & $27$  & & \multirow{2}{*}{$119$} & \multirow{2}{*}{$61$} & \multirow{2}{*}{$27$} \\
& \textsc{Yes} & $61$  & $119$ & & & & \\
\midrule
\multirow{2}{*}{TwoTurn-Prose-PC}
& \textsc{No}  & $911$ & $71$  & & \multirow{2}{*}{$128$} & \multirow{2}{*}{$52$} & \multirow{2}{*}{$71$} \\
& \textsc{Yes} & $52$  & $128$ & & & & \\
\midrule
\multirow{2}{*}{Structured Thinking}
& \textsc{No}  & $962$ & $20$  & & \multirow{2}{*}{$\mathbf{152}$} & \multirow{2}{*}{$\mathbf{28}$} & \multirow{2}{*}{$\mathbf{20}$} \\
& \textsc{Yes} & $28$  & $152$ & & & & \\
\bottomrule
\end{tabular}
\end{table}

\subsection{API configuration sensitivity}
\label{app:api-config}

On API-backed reasoning models, strict \texttt{tool\_choice="required"} can change the reasoning regime. In our initial GPT-5.4-mini run, name-forcing the \texttt{CausalAnalysis} tool reduced F$_1$ to $74.7$ ($-10.7$\,pp vs.\ the $85.4$ baseline): median reasoning tokens on $n{=}6$ fell from roughly $5000$ to roughly $800$, and $54\%$ of emitted CPDAGs had zero directed edges. Switching to \texttt{tool\_choice="auto"} plus a prompt-level instruction to call the tool preserved the same schema while restoring the reasoning budget (roughly $5600$ tokens on $n{=}6$), raising the directed-edge rate from $42\%$ to $96\%$ and recovering the reported $88.6$ F$_1$. The two configurations differ only in this API flag and prompt-level instruction.

GPT-5.4-mini has since been retired, so we repeated the comparison on a current snapshot (\texttt{gpt-5.6-terra}) with an identical schema, interface, and instance set ($n{=}1162$ per arm). Fail-closed accuracy is $90.79\%$ with \texttt{tool\_choice="auto"} and $91.05\%$ with name-forcing ($-0.26$\,pp for auto minus forcing; paired bootstrap 95\% CI $[-2.32,+1.72]$; exact McNemar $p=0.867$), with the same Turn-2 coverage in both arms ($1098/1162$). Forcing the tool has no detectable effect on this snapshot.

\section{Full-ID seed robustness}
\label{app:seed-robustness}

Because the local Qwen runs use stochastic decoding, we rerun the principal Qwen3.5-27B ID comparison on three seeds with the dataset hash, model revision (\texttt{fc05daec}), and decoding settings pinned, and with both methods run on the same seeds. Table~\ref{tab:seed-robustness} reports the full ID split ($n{=}1162$). On seed $42$, it reproduces the gap of the single run in Table~\ref{tab:main} ($+12.84$ versus $+13.35$\,pp).

\begin{table}[ht]
\caption{\textbf{Seed-matched rerun for Qwen3.5-27B on Corr2Cause ID ($n{=}1162$).} Values are percentages. ``Only S / only B'' counts items answered correctly by Structured Thinking only and by the baseline only; $p$ is the exact McNemar test on these discordant pairs. Means are followed by sample SDs over seeds.}
\label{tab:seed-robustness}
\centering
\footnotesize
\setlength{\tabcolsep}{3.5pt}
\begin{tabular}{llrrrrcc}
\toprule
\textbf{Seed} & \textbf{Method} & \textbf{F$_1$(Yes)} & \textbf{Acc.} & \textbf{Prec.} & \textbf{Rec.} & \textbf{Only S / only B} & \textbf{$p$} \\
\midrule
42 & Baseline & $74.56$ & $92.60$ & $79.75$ & $70.00$ & \multirow{2}{*}{$60$ / $19$} & \multirow{2}{*}{$4.2\times10^{-6}$} \\
42 & Structured & $87.39$ & $96.13$ & $88.14$ & $86.67$ & & \\
\midrule
52 & Baseline & $80.95$ & $94.49$ & $87.18$ & $75.56$ & \multirow{2}{*}{$32$ / $22$} & \multirow{2}{*}{$0.22$} \\
52 & Structured & $84.92$ & $95.35$ & $85.39$ & $84.44$ & & \\
\midrule
62 & Baseline & $76.92$ & $93.55$ & $86.21$ & $69.44$ & \multirow{2}{*}{$49$ / $21$} & \multirow{2}{*}{$1.1\times10^{-3}$} \\
62 & Structured & $87.12$ & $95.96$ & $85.95$ & $88.33$ & & \\
\midrule
\multicolumn{2}{l}{Baseline, mean} & $77.48 \pm 3.23$ & $93.55 \pm 0.95$ & $84.38 \pm 4.04$ & $71.67 \pm 3.38$ & & \\
\multicolumn{2}{l}{Structured, mean} & $86.48 \pm 1.36$ & $95.81 \pm 0.41$ & $86.49 \pm 1.45$ & $86.48 \pm 1.95$ & & \\
\multicolumn{2}{l}{Paired difference} & $+9.00 \pm 4.56$ & $+2.27 \pm 1.34$ & $+2.11 \pm 5.49$ & $+14.81 \pm 5.25$ & & \\
\bottomrule
\end{tabular}
\end{table}

Recall rises on every seed, by $9$ to $19$\,pp. Precision rises by $8.4$\,pp on seed $42$ and falls by $1.8$ and $0.3$\,pp on seeds $52$ and $62$. The baseline's run-to-run variation is concentrated on large positive instances: $111$ items receive different baseline predictions across the three seeds, all with five or more variables ($79$ with six) and $67$ of them positive, against $48$ items for Structured Thinking. These are also the instances where Structured Thinking helps most (Figure~\ref{fig:complexity}).

\section{Larger-graph stress tests}
\label{app:larger-graph}

\begin{table}[ht]
\caption{\textbf{Pilot larger-graph stress tests.} Values are percentages except $N$ and \#Yes. These single-seed rows use graph families outside the original Corr2Cause $n{\le}6$ range.}
\label{tab:larger-graph}
\centering
\scriptsize
\begin{tabular}{lllrrrrrr}
\toprule
\textbf{Setting} & \textbf{Model} & \textbf{Method} & \textbf{N} & \textbf{\#Yes} & \textbf{Prec.} & \textbf{Rec.} & \textbf{Acc.} & \textbf{F$_1$} \\
\midrule
Random $n{=}7$ pilot & Qwen3.5-27B & Baseline   & $300$ & $150$ & $67.50$ & $18.00$ & $54.67$ & $28.42$ \\
Random $n{=}7$ pilot & Qwen3.5-27B & Structured & $300$ & $150$ & $97.74$ & $86.67$ & $92.33$ & $\mathbf{91.87}$ \\
Random $n{=}7$ pilot & Qwen3.5-9B  & Baseline   & $300$ & $150$ & $75.76$ & $16.67$ & $55.67$ & $27.32$ \\
Random $n{=}7$ pilot & Qwen3.5-9B  & Structured & $300$ & $150$ & $70.37$ & $25.33$ & $57.33$ & $\mathbf{37.25}$ \\
ALARM-derived $n{=}7$ & Qwen3.5-27B & Baseline   & $300$ & $150$ & $43.75$ & $9.33$  & $48.67$ & $15.38$ \\
ALARM-derived $n{=}7$ & Qwen3.5-27B & Structured & $300$ & $150$ & $97.60$ & $81.33$ & $89.67$ & $\mathbf{88.73}$ \\
ALARM-derived $n{=}7$ & Qwen3.5-9B  & Baseline   & $300$ & $150$ & $47.83$ & $14.67$ & $49.33$ & $22.45$ \\
ALARM-derived $n{=}7$ & Qwen3.5-9B  & Structured & $300$ & $150$ & $65.38$ & $22.67$ & $55.33$ & $\mathbf{33.66}$ \\
ASIA-derived $n{=}8$  & Qwen3.5-27B & Baseline   & $168$ & $20$  & $8.70$  & $10.00$ & $76.79$ & $9.30$ \\
ASIA-derived $n{=}8$  & Qwen3.5-27B & Structured & $168$ & $20$  & $60.00$ & $60.00$ & $90.48$ & $\mathbf{60.00}$ \\
ASIA-derived $n{=}8$  & Qwen3.5-9B  & Baseline   & $40$  & $20$  & $100.00$& $5.00$  & $52.50$ & $9.52$ \\
ASIA-derived $n{=}8$  & Qwen3.5-9B  & Structured & $40$  & $20$  & $100.00$& $20.00$ & $60.00$ & $\mathbf{33.33}$ \\
\bottomrule
\end{tabular}
\end{table}

The 27B gains are large on all three stress settings, while the 9B rows remain positive but capacity-limited. The ASIA-derived 9B row is small ($N{=}40$). The source graph families are used to generate larger CPDAG-query instances; source-network names and semantic variable names are withheld from the model, and prompts use anonymized variable labels. Seed metadata and confusion matrices are included in the released evaluation files.

\section{Scale interaction diagnostic}
\label{app:scale-thinking}

\begin{table}[ht]
\caption{\textbf{Qwen3.5-9B factorial} ($n{=}1162$, ID split). Thinking is net-harmful to Structured Thinking at 9\,B, reversing its Qwen3.5-27B behaviour (Table~\ref{tab:ablation}).}
\label{tab:scale}
\centering
\small
\begin{tabular}{lcc}
\toprule
\textbf{Configuration} & \textbf{F$_1$(Yes)} & \textbf{Acc.} \\
\midrule
Structured Thinking (full)               & $59.56$ & $88.90$ \\
\hspace{1em}--thinking                   & $\mathbf{64.46}$ & $\mathbf{89.85}$ \\
\hspace{1em}--constrained decoding       & $45.85$ & $85.97$ \\
\midrule
BaselineEnhanced                         & $39.18$ & $84.77$ \\
\hspace{1em}--thinking                   & $45.58$ & $86.23$ \\
\bottomrule
\end{tabular}
\end{table}

Thinking helps at 27\,B but hurts at 9\,B: disabling it improves Structured Thinking by $+4.9$\,pp and the baseline by $+6.4$\,pp. Constrained decoding shows the opposite pattern, helping more at 9\,B ($13.7$\,pp) than at 27\,B ($9.5$\,pp). This suggests a practical calibration rule: enable thinking only when a validation split shows that the model's trace is reliable at the target scale.

\section{Graph validity audit}
\label{app:validator}

The validator works offline on each candidate summary. It rejects duplicate or contradictory entries and checks that the directed and undirected edges partition the skeleton. It then requires a DAG extension: an acyclic orientation of the undirected edges whose colliders are exactly the emitted v-structures. Finally, it orients those colliders on the skeleton, closes the result under Meek's rules R1--R3, which yields the CPDAG of the equivalence class, and requires this CPDAG to match the emitted directed and undirected edges exactly; this rejects both missing and superfluous orientations. A missing or unparsable summary counts as invalid (fail-closed). The validator does not interpret separating-set tokens and does not check the graph against the premise. All $1162$ reference CPDAGs and all $576$ graphs of the counterfactual probe pass it. A weaker check that keeps the emitted orientations fixed would accept over-oriented graphs (for example $A\to B\to C$ in place of $A - B - C$) and raise each validity rate by $2.5$ to $3.3$ points, without changing the comparison between runs.

\begin{table}[ht]
\caption{\textbf{Graph-theoretic validity of candidate summaries} on the full ID split ($n{=}1162$, Qwen3.5-27B). ``Usable'' means the summary parses and passes the schema; ``valid'' means it is a graph-validated CPDAG.}
\label{tab:validity}
\centering
\small
\begin{tabular}{lccc}
\toprule
\textbf{Run} & \textbf{Usable} & \textbf{Valid (all records)} & \textbf{Valid $|$ usable} \\
\midrule
Structured Thinking, constrained decoding & $97.2$ & $76.8$ & $79.0$ \\
Structured Thinking, no hard constraint & $74.5$ & $60.4$ & $81.1$ \\
Structured Thinking, independent run & $94.9$ & $81.3$ & $85.7$ \\
\bottomrule
\end{tabular}
\end{table}

The first two rows form the matched comparison discussed in \S\ref{sec:ablation}. On the $945$ graph-validated summaries of the independent run, the graph implies the official label in $930$ cases, and Turn~2 agrees with its graph in $933$.

\section{Probe details}
\label{app:scramble}

\paragraph{Counterfactual construction.} From the $1162$ ID instances, $384$ admit a counterfactual CPDAG with the same variable count under which the hypothesis flips; we froze a selection of $192$ before running the probe. For every triplet, the reference, counterfactual, and serialization-control graphs are recomputed and checked by the validator of Appendix~\ref{app:validator}, together with the truth value of the hypothesis on each graph. Turn~2 uses the standard tool-result template and decoding settings. Of the $173$ flips, $96$ go from \textsc{No} to \textsc{Yes} and $77$ from \textsc{Yes} to \textsc{No}; every flip moves toward the counterfactual graph's truth. The premise-present variant reuses the $192$ triplets from an earlier run and was analyzed post hoc.

\paragraph{Reference-CPDAG decomposition.} The reference CPDAG is built by a symbolic solver that parses the premise, runs the PC algorithm, enumerates the Markov equivalence class, and evaluates the hypothesis. On the same frozen $128$ instances, with the premise present, candidate graphs are usable for $117$, $118$, and $122$ instances on seeds $42$, $52$, and $62$. On these paired instances, accuracy with the candidate versus the reference graph is $96.58$ versus $96.58\%$, $93.22$ versus $97.46\%$, and $97.54$ versus $97.54\%$. Only seed $62$ has the full $2\times2$ design of Table~\ref{tab:probes}b.

\paragraph{Destructive scrambles.} Table~\ref{tab:scramble} reports full-split probes that corrupt fields of the emitted graph in Turn~2. They show strong dependence on graph content, but a scrambled graph need not be a valid CPDAG, which is why the counterfactual probe above is our primary evidence.

\begin{table}[ht]
\caption{\textbf{Destructive Turn-2 probes on Qwen3.5-27B ID} (single run of Table~\ref{tab:main}).}
\label{tab:scramble}
\centering
\small
\begin{tabular}{lccr}
\toprule
\textbf{Turn-2 condition} & \textbf{F$_1$(Yes)} & \textbf{Acc.} & \textbf{$\Delta$F$_1$} \\
\midrule
Full Structured (unmodified)                          & $86.36$ & $95.87$ & --- \\
no-premise (drop $P$ in Turn~2)                       & $82.56$ & $94.84$ & $-3.80$ \\
scrambled-skeleton only                               & $80.11$ & $93.98$ & $-6.25$ \\
scrambled-v-structures only                           & $80.00$ & $94.06$ & $-6.36$ \\
scrambled-directed only                               & $80.70$ & $94.32$ & $-5.66$ \\
full-scramble (all of skeleton, directed, undirected) & $74.32$ & $92.69$ & $-12.04$ \\
\bottomrule
\end{tabular}
\end{table}

\section{Cross-task transfer}
\label{app:transfer}

Both studies use \texttt{gpt-5.6-terra} and \texttt{gpt-5.6-luna} with high reasoning effort, identical frozen item sets across arms and snapshots, and a fail-closed denominator in which any protocol failure counts as an error. Turn~1 sees the story but not the query, and Turn~2 sees the emitted intermediate and the query but not the story. \emph{Direct} answers from the story in one call; \emph{Prose} restates the explicit facts as text; \emph{Structured} emits a typed relation graph through a schema-enforced tool call; \emph{Gold} gives Turn~2 the reference graph.

\paragraph{StepGame.} We use the corrected-noise release of StepGame~\citep{li2024stepgame} (MIT license) and select $360$ test items by hash, stratified over $k\in\{4,6,8,10\}$ reasoning hops and the nine relation labels. We selected StepGame after the CLUTRR result below was known and fixed its data, arms, primary comparison (Structured minus Prose), and statistical thresholds before the first full run.

\paragraph{CLUTRR.} We use $320$ frozen CLUTRR~\citep{sinha2019clutrr} items with the same four arms. Structured falls below Prose on both snapshots, and even Gold reaches only $34.06\%$: most wrong Gold answers predict no relation, so composing kinship relations from the graph alone is itself a bottleneck.

\begin{table}[ht]
\caption{\textbf{Cross-task accuracy (\%)} on two GPT-5.6 snapshots. $p$ is the exact McNemar test for Structured versus Prose.}
\label{tab:transfer}
\centering
\small
\begin{tabular}{llccccc}
\toprule
\textbf{Task} & \textbf{Snapshot} & \textbf{Direct} & \textbf{Prose} & \textbf{Structured} & \textbf{Gold} & \textbf{$p$} \\
\midrule
\multirow{2}{*}{StepGame ($n{=}360$)} & terra & $99.72$ & $70.56$ & $\mathbf{83.06}$ & $94.17$ & $1.9\times10^{-6}$ \\
 & luna & $99.44$ & $19.44$ & $\mathbf{71.39}$ & $98.06$ & $3.3\times10^{-46}$ \\
\midrule
\multirow{2}{*}{CLUTRR ($n{=}320$)} & terra & $69.06$ & $\mathbf{70.63}$ & $17.81$ & $34.06$ & $6.8\times10^{-42}$ \\
 & luna & $68.44$ & $\mathbf{55.00}$ & $22.19$ & $34.06$ & $1.9\times10^{-19}$ \\
\bottomrule
\end{tabular}
\end{table}

\section{Prompt templates}
\label{app:prompts}

We reproduce the main prompts for Structured Thinking and the direct PC baseline, as found in \texttt{src/prompts.py}. Each prompt is a Python string constant; the Python-level newlines below correspond exactly to the newlines passed to the model. Additional control variants, including the run reported here as \textsc{TwoTurn-Prose-PC} and \textsc{SingleTurn-StructPrompt}, are recorded in the released experiment metadata.

\subsection{Structured Thinking system prompt (Sstruct, Turn~1)}
\label{app:prompts:struct}
Source: \texttt{src/prompts.py:131--162} (constant \texttt{SYSTEM\_STRUCTURED\_V2}).

\begin{footnotesize}
\begin{verbatim}
You are an expert in causal inference and the PC algorithm for
causal discovery from observational data.

Given correlational premises (marginal dependencies and conditional
independencies) and a causal hypothesis, analyze the causal structure
using the PC algorithm:

## Step 1: Determine the Skeleton
- Start with a complete undirected graph over all variables.
- For each conditional independence X _||_ Y | S, REMOVE edge X-Y.
  Record S as the separating set sep(X,Y).
- CRITICAL: If two variables are adjacent in a faithful DAG, they
  CANNOT be conditionally independent given ANY subset. So any stated
  CI X _||_ Y | S proves there is NO direct edge X-Y.
- Marginal correlation does NOT imply adjacency -- variables can
  correlate through indirect paths.

## Step 2: Identify V-Structures (Colliders)
- For each non-adjacent pair (X, Y) with common neighbor Z in
  the skeleton:
  - If Z is NOT in sep(X,Y) -> orient as X -> Z <- Y (v-structure)
  - If Z IS in sep(X,Y) -> NOT a v-structure, do not orient

## Step 3: Apply Meek's Rules (iterate until no changes)
- R1: X -> Y -- Z, X and Z non-adjacent => orient Y -> Z
- R2: X -> Y -> Z, X -- Z => orient X -> Z
- R3: X -- Z, Y -> Z, W -> Z, X adj Y, X adj W, Y and W
  non-adjacent => orient X -> Z

## Step 4: Record your CPDAG using the CausalAnalysis tool

Important principles:
- Closed-world assumption: only the stated independencies hold.
  Any independence not listed is assumed to NOT hold.
- A hypothesis is deducible iff it holds in ALL DAGs of the
  Markov equivalence class (all valid completions of the CPDAG).

End your final answer with exactly "Answer: Yes" or "Answer: No".
\end{verbatim}
\end{footnotesize}

Although this legacy Turn-1 system prompt contains an answer-format sentence, the constrained decoder accepts only the \texttt{CausalAnalysis} tool call in Turn~1; final answers are generated and extracted only in Turn~2.

\subsection{Turn-2 tool-result template}
\label{app:prompts:tool}
Source: \texttt{src/prompts.py:166--187} (function \texttt{make\_tool\_result\_v2}).
After Turn~1 emits a \texttt{CausalAnalysis} object $g$, the tool-result message
fed back in Turn~2 is formatted as:

\begin{footnotesize}
\begin{verbatim}
Causal analysis recorded. Your CPDAG summary:
  Variables: <variables>
  Skeleton: <skeleton>
  Non-adjacent: <non_adjacent>
  V-structures: <v_structures>
  Directed: <directed>
  Undirected: <undirected>

<DOT visualization>

Now evaluate the hypothesis against this CPDAG:
1. Check if the relevant edge exists in the skeleton.
2. Check if its direction is determined in the directed edges.
3. For undetermined edges: enumerate ALL valid DAG completions
   and verify the hypothesis holds in EVERY one.
4. For indirect causation claims: check if a directed path
   exists in ALL valid completions.
5. Use systematic case analysis -- do NOT simply conclude
   'direction is unknown, therefore No'.
\end{verbatim}
\end{footnotesize}

\subsection{BaselineEnhanced system prompt (Sbase)}
\label{app:prompts:baseline}
Source: \texttt{src/prompts.py:115--118} (constant \texttt{SYSTEM\_BASELINE\_ENHANCED}),
which concatenates the \texttt{ENHANCED\_REASONING} block (\texttt{src/prompts.py:12--39})
with a terminal answer-format instruction.

\begin{footnotesize}
\begin{verbatim}
You are an expert in causal inference, graphical causal models, and
causal discovery from observational data.

When given correlational premises (marginal and conditional
independencies) and a causal hypothesis, apply the following
reasoning framework:

1. IDENTIFY the skeleton: which pairs of variables are adjacent
   (connected) based on marginal and conditional dependencies.
2. ORIENT v-structures (colliders): if X and Y are both adjacent
   to Z but not to each other, orient as X -> Z <- Y.
3. APPLY orientation rules (Meek's rules) to propagate edge
   directions where forced by acyclicity and existing orientations.
4. DETERMINE the Markov equivalence class: identify which edges
   have determined directions and which remain ambiguous.
5. EVALUATE the hypothesis: a causal claim can be DEDUCED if and
   only if it holds in ALL DAGs within the Markov equivalence class
   consistent with the observed independencies.

Important principles:
- Assume the stated statistical relations are exhaustive (the
  closed-world assumption): any independence not listed is a dependence.
- Conditional independence X _||_ Y | Z constrains the graph structure.
- Some causal directions CAN be determined from observational data
  (especially via v-structures and orientation propagation).
- A hypothesis is deducible if every faithful DAG consistent with
  the stated independencies entails that hypothesis.
- A hypothesis is NOT deducible if there exists at least one
  faithful DAG consistent with the premises where it does not hold.

End your response with exactly "Answer: Yes" or "Answer: No".
\end{verbatim}
\end{footnotesize}

The baseline is a strong PC-instruction baseline rather than a plain
chain-of-thought baseline: it covers skeleton construction, v-structures, Meek's
rules, Markov equivalence classes, and the closed-world assumption. It is not
token-identical to $S_{\mathrm{struct}}$; in particular, its collider rule omits
the condition that $Z$ is not in the separating set of $X$ and $Y$.

\subsection{Control prompt variants}
\label{app:prompts:controls}

\textsc{TwoTurn-Prose-PC} is a two-turn prose control with the same PC scaffold as \textsc{BaselineEnhanced}. Turn~1 uses the baseline's skeleton, v-structure, Meek-rule, closed-world, and MEC-universal-quantifier instructions, but asks for free-text analysis and explicitly withholds the final yes/no answer. Turn~2 returns that free-text reasoning and asks for the final answer. This control uses no tool call and no schema-constrained typed graph object. Its prompt template is:

\begin{footnotesize}
\begin{verbatim}
Turn 1:
<BaselineEnhanced system prompt above, with the final
answer-format instruction replaced by:>

Do not give the final yes/no answer yet. Write your causal analysis in
free text, following the PC-algorithm scaffold above.

Turn 2:
Here is your previous analysis:
<turn1_analysis>

Now answer the original causal hypothesis. End your response with
exactly "Answer: Yes" or "Answer: No".
\end{verbatim}
\end{footnotesize}

\textsc{SingleTurn-StructPrompt} reuses the structured CPDAG-oriented instructions in a single-turn direct-answer setting, but disables the tool call and schema-constrained graph output. These controls help distinguish graph-object externalization from a PC-scaffolded prose extra turn and graph vocabulary alone.

\section{Schema and constrained-decoding grammar}
\label{app:schema}

\subsection{\texttt{CausalAnalysis} schema}
Source: \texttt{src/schema.py:194--241}. The schema is defined with Pydantic
and converted to JSON Schema via \texttt{CausalAnalysis.model\_json\_schema()}.

\begin{footnotesize}
\begin{verbatim}
class CausalAnalysis(BaseModel):
    """Structured causal analysis following the PC algorithm.

    Determine the causal skeleton, identify v-structures, apply Meek's
    rules, and record the resulting CPDAG.
    """
    variables:    list[str]   # e.g. ['A', 'B', 'C', 'D']
    skeleton:     list[str]   # adjacent pairs, e.g. ['A-B', 'C-D']
    non_adjacent: list[str]   # non-adj pairs w/ sep set, e.g. ['A-C|B']
    v_structures: list[str]   # e.g. ['A->E<-C']
    directed:     list[str]   # oriented edges, e.g. ['A->E', 'E->F']
    undirected:   list[str]   # e.g. ['D-E', 'A-F']
\end{verbatim}
\end{footnotesize}

\subsection{Regex construction (Qwen / Mistral)}
Source: \texttt{src/schema.py:248--286}. The regex derived from the JSON
schema via \texttt{outlines\_core.json\_schema.build\_regex\_from\_schema}
is wrapped in a model-specific envelope.

\begin{footnotesize}
\begin{verbatim}
# Qwen family (Qwen3.5-27B, Qwen3.6-27B, Qwen3.5-9B) -- chat template uses
# <think>...</think> for reasoning and <tool_call>...</tool_call>
# for tool invocation.
def build_analysis_regex(enable_thinking: bool = True) -> str:
    schema_str = json.dumps(CausalAnalysis.model_json_schema())
    args_regex = build_regex_from_schema(schema_str)
    body = (
        r'\{"name":\s*"CausalAnalysis",\s*"arguments":\s*'
        + args_regex + r"\}"
    )
    tool_call = r"<tool_call>\s*" + body + r"\s*</tool_call>"
    if enable_thinking:
        return r"<think>\s*[\s\S]+?</think>\s*" + tool_call
    return tool_call

# Magistral / Mistral family -- chat template uses [THINK]...[/THINK]
# and emits the raw JSON object without a tool_call envelope.
def build_analysis_regex_mistral(enable_thinking: bool = True) -> str:
    schema_str = json.dumps(CausalAnalysis.model_json_schema())
    args_regex = build_regex_from_schema(schema_str)
    if enable_thinking:
        return r"\[THINK\][\s\S]+?\[/THINK\]\s*" + args_regex
    return args_regex
\end{verbatim}
\end{footnotesize}

For API models (\texttt{gpt-5.4-mini}) the same schema is enforced through
OpenAI tool-calling with \texttt{strict}=\texttt{true}; the reported main run
uses \texttt{tool\_choice}=\texttt{"auto"} plus a prompt-level instruction to
call the tool, because name-forcing the tool compressed reasoning
(Appendix~\ref{app:api-config}).

\section{Per-experiment tables}
\label{app:tables}

Full per-experiment results (F$_1$, accuracy, precision, recall, per-complexity F$_1$, confusion matrices) for every reported experiment in the main text and appendix are included in the released artifact as one \texttt{evaluation.json} per experiment directory (\texttt{experiments/<name>/evaluation.json}), together with audit outputs for Table~\ref{tab:cpdag-audit}. Token and wall-clock cost per experiment are similarly recorded in \texttt{experiments/<name>/usage.json} and summarized in \texttt{analysis/cost\_table.json}. We omit the verbatim per-cell tables from the appendix to stay within the page budget.

\section{Assets and licenses}
\label{app:assets}

Corr2Cause is used as released by \citet{jin2024corr2cause}. The open-weight models are accessed through public Hugging Face releases: \texttt{Qwen/Qwen3.5-27B}, \texttt{Qwen/Qwen3.6-27B}, \texttt{Qwen/Qwen3.5-9B}, and \texttt{google/gemma-4-31B-it}; revision hashes are recorded in the run metadata. The corresponding model cards or license files report Apache-2.0 licensing. GPT-5.4-mini and the GPT-5.6 snapshots are accessed through the OpenAI API under the applicable API terms; snapshot identifiers are recorded in the run metadata. StepGame is used in its corrected-noise release under the MIT license~\citep{li2024stepgame}, and CLUTRR from its official repository~\citep{sinha2019clutrr}. vLLM, XGrammar, and the schema-constrained decoding dependencies are cited in the main text; exact package versions are recorded in the released metadata. The code and data artifact is available at \url{https://github.com/wentaopoly/Structured-Thinking-Matters}; it contains experiment outputs, prompts, schemas, and analysis scripts produced for this paper. No new model weights or human-subject data are released.

\section{Failure-case gallery}
\label{app:failures}

\begin{table}[ht]
\caption{\textbf{Failure-mode distribution} on 32 hand-labelled Structured Thinking errors (Qwen3.5-27B, ID split).}
\label{tab:failures}
\centering
\small
\begin{tabular}{lcc}
\toprule
\textbf{Failure mode} & \textbf{Count} & \textbf{\%} \\
\midrule
\textsc{CPDAG\_Wrong} --- missing or spurious graph edges & $13$ & $40.6$ \\
\textsc{Semantic\_Error} --- correct graph, wrong Turn-2 readout & $8$  & $25.0$ \\
\textsc{Undirected\_MEC} --- MEC ambiguity collapsed incorrectly & $6$  & $18.8$ \\
\textsc{KG\_Invalid} --- schema-valid but logically malformed & $5$  & $15.6$ \\
\bottomrule
\end{tabular}
\end{table}

We hand-label 32 representative Structured Thinking errors selected to cover the observed failure categories; the gallery is diagnostic rather than an unbiased prevalence estimate. The records are included in the released artifact as \texttt{analysis/failure\_modes.json}, each providing the instance index, variable list, ground-truth label, model prediction, and error category. Two illustrative worked examples follow.

\paragraph{\textsc{CPDAG\_Wrong} (idx 321).} Premise involves 5 variables. Ground-truth CPDAG has $A\!-\!E$ in the skeleton as part of a v-structure; the model emitted a skeleton that omitted $A\!-\!E$, so Turn-2 lacks the necessary edge and answered \textsc{Yes} to ``$A$ is parent of $E$'' via an irrelevant alternative path.

\paragraph{\textsc{Undirected\_MEC}.} In these errors, the model treats an undirected CPDAG edge as if one orientation were forced, or checks only one valid completion of the Markov equivalence class. For a parent claim on an undirected edge $X\!-\!Y$, the correct universal-quantifier test must reject the claim whenever both $X\!\to\!Y$ and $X\!\leftarrow\!Y$ remain valid completions. These cases illustrate that schema-valid graph summaries do not guarantee a correct all-completions readout.

\section{Reproducibility}
\label{app:repro}

Every experiment can be reproduced by a single command using the released code at \url{https://github.com/wentaopoly/Structured-Thinking-Matters}. Chain scripts at the repository root orchestrate batched runs: \texttt{run\_qwen9b\_chain.sh} and \texttt{run\_qwen9b\_ablation\_chain.sh} cover the Qwen3.5-9B main and ablation runs; \texttt{run\_probe\_chain.sh} covers the full-split probes of \S\ref{sec:probes}; \texttt{run\_additional\_chain.sh} orchestrates additional diagnostic runs; \texttt{run\_ood\_all.sh} produces the Refactor-OOD results; the seed-matched Qwen3.5-27B rerun in Appendix~\ref{app:seed-robustness} is recorded with the same per-experiment metadata format. Open-weight runs use 2$\times$ GPU tensor-parallel vLLM inference with \texttt{gpu\_memory\_utilization}$=\!0.88$, and API runs use \texttt{max\_concurrency}$=\!5$. All \texttt{metadata.json} files record the exact decoding arguments (temperature, top-$p$, max length, tensor-parallel size, tokenizer mode, constrained-decoding flag), Hugging Face model revision hashes, tokenizer revisions, API model snapshot strings, and package versions for each experiment. The Corr2Cause dataset is used as released at \url{https://huggingface.co/datasets/causal-nlp/corr2cause}; Refactor-OOD and Paraphrase-OOD are its \texttt{perturbation\_by\_refactorization\_test.csv} and \texttt{perturbation\_by\_paraphrasing\_test.csv} files.

\newpage
\section*{NeurIPS Paper Checklist}

\begin{enumerate}

\item {\bf Claims}
    \item[] Question: Do the main claims made in the abstract and introduction accurately reflect the paper's contributions and scope?
    \item[] Answer: \answerYes{}
    \item[] Justification: The abstract and \S\ref{sec:intro} state the main claims: gains over a strong PC-instruction baseline across three seeds, cross-model and perturbed-split support, ablation evidence, graph-validity and counterfactual probes, and a reference-CPDAG decomposition; limitations are scoped in \S\ref{sec:limitations}.
    \item[] Guidelines:
    \begin{itemize}
        \item The answer \answerNA{} means that the abstract and introduction do not include the claims made in the paper.
        \item The abstract and/or introduction should clearly state the claims made, including the contributions made in the paper and important assumptions and limitations. A \answerNo{} or \answerNA{} answer to this question will not be perceived well by the reviewers. 
        \item The claims made should match theoretical and experimental results, and reflect how much the results can be expected to generalize to other settings. 
        \item It is fine to include aspirational goals as motivation as long as it is clear that these goals are not attained by the paper. 
    \end{itemize}

\item {\bf Limitations}
    \item[] Question: Does the paper discuss the limitations of the work performed by the authors?
    \item[] Answer: \answerYes{}
    \item[] Justification: \S\ref{sec:limitations} discusses the CPDAG-deduction scope and the observed cross-task boundary (CLUTRR), the three-seed basis and seed-dependent size of the principal Qwen3.5-27B gain, the remaining single-run perturbed-split, larger-graph, cross-model, and control rows, the share of test instances that do not yield a graph-validated CPDAG and the validator's scope, and bundled method components; backend/tool-choice sensitivity and inference cost are discussed in \S\ref{sec:cost}.
    \item[] Guidelines:
    \begin{itemize}
        \item The answer \answerNA{} means that the paper has no limitation while the answer \answerNo{} means that the paper has limitations, but those are not discussed in the paper. 
        \item The authors are encouraged to create a separate ``Limitations'' section in their paper.
        \item The paper should point out any strong assumptions and how robust the results are to violations of these assumptions (e.g., independence assumptions, noiseless settings, model well-specification, asymptotic approximations only holding locally). The authors should reflect on how these assumptions might be violated in practice and what the implications would be.
        \item The authors should reflect on the scope of the claims made, e.g., if the approach was only tested on a few datasets or with a few runs. In general, empirical results often depend on implicit assumptions, which should be articulated.
        \item The authors should reflect on the factors that influence the performance of the approach. For example, a facial recognition algorithm may perform poorly when image resolution is low or images are taken in low lighting. Or a speech-to-text system might not be used reliably to provide closed captions for online lectures because it fails to handle technical jargon.
        \item The authors should discuss the computational efficiency of the proposed algorithms and how they scale with dataset size.
        \item If applicable, the authors should discuss possible limitations of their approach to address problems of privacy and fairness.
        \item While the authors might fear that complete honesty about limitations might be used by reviewers as grounds for rejection, a worse outcome might be that reviewers discover limitations that aren't acknowledged in the paper. The authors should use their best judgment and recognize that individual actions in favor of transparency play an important role in developing norms that preserve the integrity of the community. Reviewers will be specifically instructed to not penalize honesty concerning limitations.
    \end{itemize}

\item {\bf Theory assumptions and proofs}
    \item[] Question: For each theoretical result, does the paper provide the full set of assumptions and a complete (and correct) proof?
    \item[] Answer: \answerNA{}
    \item[] Justification: The paper is empirical; it makes no new theoretical claims. The PC-algorithm and Markov-equivalence assumptions inherited from Corr2Cause~\citep{jin2024corr2cause} (faithfulness, causal Markov) are stated in \S\ref{sec:task}.
    \item[] Guidelines:
    \begin{itemize}
        \item The answer \answerNA{} means that the paper does not include theoretical results. 
        \item All the theorems, formulas, and proofs in the paper should be numbered and cross-referenced.
        \item All assumptions should be clearly stated or referenced in the statement of any theorems.
        \item The proofs can either appear in the main paper or the supplemental material, but if they appear in the supplemental material, the authors are encouraged to provide a short proof sketch to provide intuition. 
        \item Inversely, any informal proof provided in the core of the paper should be complemented by formal proofs provided in appendix or supplemental material.
        \item Theorems and Lemmas that the proof relies upon should be properly referenced. 
    \end{itemize}

    \item {\bf Experimental result reproducibility}
    \item[] Question: Does the paper fully disclose all the information needed to reproduce the main experimental results of the paper to the extent that it affects the main claims and/or conclusions of the paper (regardless of whether the code and data are provided or not)?
    \item[] Answer: \answerYes{}
    \item[] Justification: \S\ref{sec:setup} reports backend, decoding parameters, exact model IDs, datasets, and significance-test protocol; Appendix~\ref{app:prompts} gives the main Structured Thinking and direct PC-baseline prompts verbatim and describes the PC-scaffolded prose and single-turn control variants; Appendix~\ref{app:schema} gives the Pydantic schema and regex construction; Appendix~\ref{app:repro} gives the repository URL, reproduction commands, and released artifacts, with model revisions, API snapshots, larger-graph stress-test metadata, and all control prompts recorded in experiment metadata.
    \item[] Guidelines:
    \begin{itemize}
        \item The answer \answerNA{} means that the paper does not include experiments.
        \item If the paper includes experiments, a \answerNo{} answer to this question will not be perceived well by the reviewers: Making the paper reproducible is important, regardless of whether the code and data are provided or not.
        \item If the contribution is a dataset and\slash or model, the authors should describe the steps taken to make their results reproducible or verifiable. 
        \item Depending on the contribution, reproducibility can be accomplished in various ways. For example, if the contribution is a novel architecture, describing the architecture fully might suffice, or if the contribution is a specific model and empirical evaluation, it may be necessary to either make it possible for others to replicate the model with the same dataset, or provide access to the model. In general. releasing code and data is often one good way to accomplish this, but reproducibility can also be provided via detailed instructions for how to replicate the results, access to a hosted model (e.g., in the case of a large language model), releasing of a model checkpoint, or other means that are appropriate to the research performed.
        \item While NeurIPS does not require releasing code, the conference does require all submissions to provide some reasonable avenue for reproducibility, which may depend on the nature of the contribution. For example
        \begin{enumerate}
            \item If the contribution is primarily a new algorithm, the paper should make it clear how to reproduce that algorithm.
            \item If the contribution is primarily a new model architecture, the paper should describe the architecture clearly and fully.
            \item If the contribution is a new model (e.g., a large language model), then there should either be a way to access this model for reproducing the results or a way to reproduce the model (e.g., with an open-source dataset or instructions for how to construct the dataset).
            \item We recognize that reproducibility may be tricky in some cases, in which case authors are welcome to describe the particular way they provide for reproducibility. In the case of closed-source models, it may be that access to the model is limited in some way (e.g., to registered users), but it should be possible for other researchers to have some path to reproducing or verifying the results.
        \end{enumerate}
    \end{itemize}

\item {\bf Open access to data and code}
    \item[] Question: Does the paper provide open access to the data and code, with sufficient instructions to faithfully reproduce the main experimental results, as described in supplemental material?
    \item[] Answer: \answerYes{}
    \item[] Justification: Code and data are available at \url{https://github.com/wentaopoly/Structured-Thinking-Matters}, including prompts, schema, the full \texttt{CausalAnalysis} specification, chain scripts (\texttt{run\_qwen9b\_chain.sh}, \texttt{run\_probe\_chain.sh}, \texttt{run\_additional\_chain.sh}), evaluation/usage/metadata JSON artefacts, and the 32-case failure set; the Corr2Cause dataset is public~\citep{jin2024corr2cause}.
    \item[] Guidelines:
    \begin{itemize}
        \item The answer \answerNA{} means that paper does not include experiments requiring code.
        \item Please see the NeurIPS code and data submission guidelines (\url{https://neurips.cc/public/guides/CodeSubmissionPolicy}) for more details.
        \item While we encourage the release of code and data, we understand that this might not be possible, so \answerNo{} is an acceptable answer. Papers cannot be rejected simply for not including code, unless this is central to the contribution (e.g., for a new open-source benchmark).
        \item The instructions should contain the exact command and environment needed to run to reproduce the results. See the NeurIPS code and data submission guidelines (\url{https://neurips.cc/public/guides/CodeSubmissionPolicy}) for more details.
        \item The authors should provide instructions on data access and preparation, including how to access the raw data, preprocessed data, intermediate data, and generated data, etc.
        \item The authors should provide scripts to reproduce all experimental results for the new proposed method and baselines. If only a subset of experiments are reproducible, they should state which ones are omitted from the script and why.
        \item At submission time, to preserve anonymity, the authors should release anonymized versions (if applicable).
        \item Providing as much information as possible in supplemental material (appended to the paper) is recommended, but including URLs to data and code is permitted.
    \end{itemize}

\item {\bf Experimental setting/details}
    \item[] Question: Does the paper specify all the training and test details (e.g., data splits, hyperparameters, how they were chosen, type of optimizer) necessary to understand the results?
    \item[] Answer: \answerYes{}
    \item[] Justification: No training; all results are inference-only. Decoding hyperparameters ($T{=}0.6$, top-$p{=}0.95$, max length $32768$), backends (vLLM for open-weight, OpenAI Responses API for \texttt{gpt-5.4-mini}), constrained-decoding implementations, split sizes, model revision hashes, and API snapshots are reported in \S\ref{sec:setup} and in the per-experiment \texttt{metadata.json} files.
    \item[] Guidelines:
    \begin{itemize}
        \item The answer \answerNA{} means that the paper does not include experiments.
        \item The experimental setting should be presented in the core of the paper to a level of detail that is necessary to appreciate the results and make sense of them.
        \item The full details can be provided either with the code, in appendix, or as supplemental material.
    \end{itemize}

\item {\bf Experiment statistical significance}
    \item[] Question: Does the paper report error bars suitably and correctly defined or other appropriate information about the statistical significance of the experiments?
    \item[] Answer: \answerYes{}
    \item[] Justification: The principal Qwen3.5-27B ID comparison is a seed-matched rerun on three seeds: \S\ref{sec:results} and Appendix~\ref{app:seed-robustness} report per-seed exact McNemar tests on the paired $n{=}1162$ items and the mean $\pm$ sample standard deviation across seeds, which captures run-to-run variation from stochastic decoding. Probes and cross-task studies report exact McNemar tests on paired items, and the API tool-choice control a paired-bootstrap 95\% CI. \S\ref{sec:limitations} states that the perturbed splits, larger-graph stress tests, cross-model rows, and controls are single runs.
    \item[] Guidelines:
    \begin{itemize}
        \item The answer \answerNA{} means that the paper does not include experiments.
        \item The authors should answer \answerYes{} if the results are accompanied by error bars, confidence intervals, or statistical significance tests, at least for the experiments that support the main claims of the paper.
        \item The factors of variability that the error bars are capturing should be clearly stated (for example, train/test split, initialization, random drawing of some parameter, or overall run with given experimental conditions).
        \item The method for calculating the error bars should be explained (closed form formula, call to a library function, bootstrap, etc.)
        \item The assumptions made should be given (e.g., Normally distributed errors).
        \item It should be clear whether the error bar is the standard deviation or the standard error of the mean.
        \item It is OK to report 1-sigma error bars, but one should state it. The authors should preferably report a 2-sigma error bar than state that they have a 96\% CI, if the hypothesis of Normality of errors is not verified.
        \item For asymmetric distributions, the authors should be careful not to show in tables or figures symmetric error bars that would yield results that are out of range (e.g., negative error rates).
        \item If error bars are reported in tables or plots, the authors should explain in the text how they were calculated and reference the corresponding figures or tables in the text.
    \end{itemize}

\item {\bf Experiments compute resources}
    \item[] Question: For each experiment, does the paper provide sufficient information on the computer resources (type of compute workers, memory, time of execution) needed to reproduce the experiments?
    \item[] Answer: \answerYes{}
    \item[] Justification: Cost is summarized in \S\ref{sec:cost} and derived from the committed \texttt{usage.json} files; open-weight runs use 2$\times$ GPU tensor-parallel vLLM inference at \texttt{gpu\_memory\_utilization}$=\!0.88$; API runs use the OpenAI Responses API with \texttt{max\_concurrency}$=\!5$. The seed-matched rerun, probes, and reference-CPDAG studies ran on NVIDIA H100, A100, and RTX 6000 Ada GPUs, and the cross-task studies used the OpenAI API.
    \item[] Guidelines:
    \begin{itemize}
        \item The answer \answerNA{} means that the paper does not include experiments.
        \item The paper should indicate the type of compute workers CPU or GPU, internal cluster, or cloud provider, including relevant memory and storage.
        \item The paper should provide the amount of compute required for each of the individual experimental runs as well as estimate the total compute. 
        \item The paper should disclose whether the full research project required more compute than the experiments reported in the paper (e.g., preliminary or failed experiments that didn't make it into the paper). 
    \end{itemize}
    
\item {\bf Code of ethics}
    \item[] Question: Does the research conducted in the paper conform, in every respect, with the NeurIPS Code of Ethics \url{https://neurips.cc/public/EthicsGuidelines}?
    \item[] Answer: \answerYes{}
    \item[] Justification: The research uses public benchmarks and released model weights or APIs, involves no human subjects, and makes no privacy-sensitive data accesses.
    \item[] Guidelines:
    \begin{itemize}
        \item The answer \answerNA{} means that the authors have not reviewed the NeurIPS Code of Ethics.
        \item If the authors answer \answerNo, they should explain the special circumstances that require a deviation from the Code of Ethics.
        \item The authors should make sure to preserve anonymity (e.g., if there is a special consideration due to laws or regulations in their jurisdiction).
    \end{itemize}

\item {\bf Broader impacts}
    \item[] Question: Does the paper discuss both potential positive societal impacts and negative societal impacts of the work performed?
    \item[] Answer: \answerYes{}
    \item[] Justification: The method is an evaluation/prompting technique for causal-deduction benchmarks. Potential positive impact is better auditing of LLM causal reasoning; potential negative impact is over-trust if structured outputs are mistaken for certified causal discovery. \S\ref{sec:limitations} reports how often test instances do not yield a valid CPDAG, and the method should not be used as a substitute for a causal solver.
    \item[] Guidelines:
    \begin{itemize}
        \item The answer \answerNA{} means that there is no societal impact of the work performed.
        \item If the authors answer \answerNA{} or \answerNo, they should explain why their work has no societal impact or why the paper does not address societal impact.
        \item Examples of negative societal impacts include potential malicious or unintended uses (e.g., disinformation, generating fake profiles, surveillance), fairness considerations (e.g., deployment of technologies that could make decisions that unfairly impact specific groups), privacy considerations, and security considerations.
        \item The conference expects that many papers will be foundational research and not tied to particular applications, let alone deployments. However, if there is a direct path to any negative applications, the authors should point it out. For example, it is legitimate to point out that an improvement in the quality of generative models could be used to generate Deepfakes for disinformation. On the other hand, it is not needed to point out that a generic algorithm for optimizing neural networks could enable people to train models that generate Deepfakes faster.
        \item The authors should consider possible harms that could arise when the technology is being used as intended and functioning correctly, harms that could arise when the technology is being used as intended but gives incorrect results, and harms following from (intentional or unintentional) misuse of the technology.
        \item If there are negative societal impacts, the authors could also discuss possible mitigation strategies (e.g., gated release of models, providing defenses in addition to attacks, mechanisms for monitoring misuse, mechanisms to monitor how a system learns from feedback over time, improving the efficiency and accessibility of ML).
    \end{itemize}
    
\item {\bf Safeguards}
    \item[] Question: Does the paper describe safeguards that have been put in place for responsible release of data or models that have a high risk for misuse (e.g., pre-trained language models, image generators, or scraped datasets)?
    \item[] Answer: \answerNA{}
    \item[] Justification: No new models or data are released; the released artefact is inference-time code using public models and the public Corr2Cause dataset.
    \item[] Guidelines:
    \begin{itemize}
        \item The answer \answerNA{} means that the paper poses no such risks.
        \item Released models that have a high risk for misuse or dual-use should be released with necessary safeguards to allow for controlled use of the model, for example by requiring that users adhere to usage guidelines or restrictions to access the model or implementing safety filters. 
        \item Datasets that have been scraped from the Internet could pose safety risks. The authors should describe how they avoided releasing unsafe images.
        \item We recognize that providing effective safeguards is challenging, and many papers do not require this, but we encourage authors to take this into account and make a best faith effort.
    \end{itemize}

\item {\bf Licenses for existing assets}
    \item[] Question: Are the creators or original owners of assets (e.g., code, data, models), used in the paper, properly credited and are the license and terms of use explicitly mentioned and properly respected?
    \item[] Answer: \answerYes{}
    \item[] Justification: Corr2Cause~\citep{jin2024corr2cause}, StepGame~\citep{li2024stepgame} (MIT), and CLUTRR~\citep{sinha2019clutrr} are used under their released licenses; Qwen3.5, Qwen3.6, Gemma-4, GPT-5.4-mini, and GPT-5.6 are accessed under their respective model licenses / API terms; Outlines~\citep{willard2023outlines}, XGrammar~\citep{dong2024xgrammar}, and vLLM~\citep{kwon2023vllm} are used under their open-source licenses. Appendix~\ref{app:assets} summarizes assets and licensing.
    \item[] Guidelines:
    \begin{itemize}
        \item The answer \answerNA{} means that the paper does not use existing assets.
        \item The authors should cite the original paper that produced the code package or dataset.
        \item The authors should state which version of the asset is used and, if possible, include a URL.
        \item The name of the license (e.g., CC-BY 4.0) should be included for each asset.
        \item For scraped data from a particular source (e.g., website), the copyright and terms of service of that source should be provided.
        \item If assets are released, the license, copyright information, and terms of use in the package should be provided. For popular datasets, \url{paperswithcode.com/datasets} has curated licenses for some datasets. Their licensing guide can help determine the license of a dataset.
        \item For existing datasets that are re-packaged, both the original license and the license of the derived asset (if it has changed) should be provided.
        \item If this information is not available online, the authors are encouraged to reach out to the asset's creators.
    \end{itemize}

\item {\bf New assets}
    \item[] Question: Are new assets introduced in the paper well documented and is the documentation provided alongside the assets?
    \item[] Answer: \answerYes{}
    \item[] Justification: The released repository contains the \texttt{CausalAnalysis} schema, all prompts, the two-turn pipeline implementation (\texttt{pipeline.py}, \texttt{pipeline\_api.py}), evaluation scripts, and significance analysis (\texttt{analysis/significance.py}); a README documents usage.
    \item[] Guidelines:
    \begin{itemize}
        \item The answer \answerNA{} means that the paper does not release new assets.
        \item Researchers should communicate the details of the dataset\slash code\slash model as part of their submissions via structured templates. This includes details about training, license, limitations, etc. 
        \item The paper should discuss whether and how consent was obtained from people whose asset is used.
        \item At submission time, remember to anonymize your assets (if applicable). You can either create an anonymized URL or include an anonymized zip file.
    \end{itemize}

\item {\bf Crowdsourcing and research with human subjects}
    \item[] Question: For crowdsourcing experiments and research with human subjects, does the paper include the full text of instructions given to participants and screenshots, if applicable, as well as details about compensation (if any)?
    \item[] Answer: \answerNA{}
    \item[] Justification: The research involves no crowdsourcing or human subjects; the 32-case failure taxonomy was hand-labelled by the authors.
    \item[] Guidelines:
    \begin{itemize}
        \item The answer \answerNA{} means that the paper does not involve crowdsourcing nor research with human subjects.
        \item Including this information in the supplemental material is fine, but if the main contribution of the paper involves human subjects, then as much detail as possible should be included in the main paper. 
        \item According to the NeurIPS Code of Ethics, workers involved in data collection, curation, or other labor should be paid at least the minimum wage in the country of the data collector. 
    \end{itemize}

\item {\bf Institutional review board (IRB) approvals or equivalent for research with human subjects}
    \item[] Question: Does the paper describe potential risks incurred by study participants, whether such risks were disclosed to the subjects, and whether Institutional Review Board (IRB) approvals (or an equivalent approval/review based on the requirements of your country or institution) were obtained?
    \item[] Answer: \answerNA{}
    \item[] Justification: The research involves no human subjects.
    \item[] Guidelines:
    \begin{itemize}
        \item The answer \answerNA{} means that the paper does not involve crowdsourcing nor research with human subjects.
        \item Depending on the country in which research is conducted, IRB approval (or equivalent) may be required for any human subjects research. If you obtained IRB approval, you should clearly state this in the paper. 
        \item We recognize that the procedures for this may vary significantly between institutions and locations, and we expect authors to adhere to the NeurIPS Code of Ethics and the guidelines for their institution. 
        \item For initial submissions, do not include any information that would break anonymity (if applicable), such as the institution conducting the review.
    \end{itemize}

\item {\bf Declaration of LLM usage}
    \item[] Question: Does the paper describe the usage of LLMs if it is an important, original, or non-standard component of the core methods in this research? Note that if the LLM is used only for writing, editing, or formatting purposes and does \emph{not} impact the core methodology, scientific rigor, or originality of the research, declaration is not required.
    \item[] Answer: \answerYes{}
    \item[] Justification: LLMs are the core subject of study: the method prompts open-weight models (Qwen3.5-9B/27B, Qwen3.6-27B, Gemma-4-31B) and API models (GPT-5.4-mini, GPT-5.6) to emit a structured CPDAG-style summary and answer a causal hypothesis against it. The main prompt templates, schemas, and decoding constraints are reported in \S\ref{sec:method} and Appendices~\ref{app:prompts}--\ref{app:schema}; control prompt variants are described in Appendix~\ref{app:prompts:controls} and recorded in the released experiment metadata.
    \item[] Guidelines:
    \begin{itemize}
        \item The answer \answerNA{} means that the core method development in this research does not involve LLMs as any important, original, or non-standard components.
        \item Please refer to our LLM policy in the NeurIPS handbook for what should or should not be described.
    \end{itemize}

\end{enumerate}

\end{document}